\documentclass[letterpaper, 10pt, conference]{ieeeconf}      

\IEEEoverridecommandlockouts                              

\usepackage{graphics} 
\usepackage{epsfig} 
\usepackage{subfigure}
\usepackage{amsmath} 
\usepackage{amssymb}  
\usepackage{epstopdf}
\usepackage{lipsum}
\usepackage{lipsum,amsmath,multicol}
\usepackage{float}
\usepackage{algorithm}
\usepackage{algpseudocode}
\usepackage{wrapfig}
\usepackage{sidecap}

\usepackage{amsthm}

\usepackage{comment}
\usepackage{array}
\usepackage{glossaries}
\makeglossaries
\newcolumntype{L}[1]{>{\raggedright\let\newline\\\arraybackslash\hspace{0pt}}m{#1}}
\newcolumntype{C}[1]{>{\centering\let\newline\\\arraybackslash\hspace{0pt}}m{#1}}
\newcolumntype{R}[1]{>{\raggedleft\let\newline\\\arraybackslash\hspace{0pt}}m{#1}}
\usepackage{sidecap}
\usepackage{url}
\usepackage[T1]{fontenc}
\usepackage{xcolor}
\usepackage{tabularx}
\usepackage{multicol, blindtext}
\usepackage{amsmath}
\usepackage{multirow}

\allowdisplaybreaks

\makeatletter
\def\endthebibliography{%
  \def\@noitemerr{\@latex@warning{Empty `thebibliography' environment}}%
  \endlist
}
\makeatother

\begin{document}
%
\title{Embedded Hardware Appropriate Fast 3D Trajectory Optimization for Fixed Wing Aerial Vehicles  by Leveraging Hidden Convex Structures}
\author{Vivek Kantilal Adajania, Houman Masnavi, Fatemeh Rastgar, Karl Kruusamae, Arun Kumar Singh \thanks{Vivek is from NIT Surat, India. Rest all authors are with the Institute of Technology, University of Tartu. The work was supported in part by the European Social Fund through IT Academy program in Estonia, Estonian Centre of Excellence in IT (EXCITE) funded by the European Regional Development Fund and grants COVSG24 and PSG605 from Estonian Research Council. }
}
\maketitle



\begin{abstract}
Most commercially available fixed-wing aerial vehicles (FWV) can carry only small, lightweight computing hardware such as Jetson TX2 onboard. Solving non-linear trajectory optimization on these computing resources is computationally challenging even while considering only the kinematic motion model. Most importantly, the computation time increases sharply as the environment becomes more cluttered. In this paper, we take a step towards overcoming this bottleneck and propose a trajectory optimizer that achieves online performance on both conventional laptops/desktops and Jetson TX2 in a typical urban environment setting. Our optimizer builds on the novel insight that the seemingly non-linear trajectory optimization problem for FWV has an implicit multi-convex structure. Our optimizer exploits these computational structures by bringing together diverse concepts from Alternating Minimization, Bregman iteration, and Alternating Direction Method of Multipliers. We show that our optimizer outperforms the state-of-the-art implementation of sequential quadratic programming approach in optimal control solver ACADO in computation time and solution quality measured in terms of control and goal reaching cost.
 
\end{abstract}

\section{Introduction}

Autonomous Fixed-Wing Aerial Vehicles (FWV) are extensively used to explore and map large outdoor terrains. With an increase in commercial availability of FWVs, their application domains are expected to expand to novel challenging areas like search and rescue or surveillance in obstacle-rich urban environments and indoor spaces. To realize the full potential of FWVs, we need high-fidelity trajectory optimization algorithms that can generate optimal, collision-free trajectories from arbitrary initial states in (near) real-time. Furthermore, it is highly beneficial if these algorithms can retain online performance on light-weight embedded computers like NVIDIA Jetson-TX2 so that FWVs can perform complex on-board motion planning without requiring constant communication to a centralized server. The core computational challenge of trajectory optimization in the context of FWVs stems from the fact that even at the kinematic level, the mapping from control commands to states is highly non-linear. This in turn introduces difficult non-convexity into the problem. Existing works rely on techniques like Gradient Descent (GD) \cite{fwv_gd} or Sequential Quadratic Programming (SQP) \cite{sqp_fwv_adhikari} that at the most fundamental level, are based on first-order Taylor series expansion of the non-convex constraints and/or cost functions. Although rigorous, these techniques are not designed to leverage the niche mathematical structures in the optimization problems. For example, it is well know that GD based optimizer performs poorly on some non-convex problems with bi-convex structure \cite{li_biconvex}. As a result, without careful initialization, even the state of the art implementation of GD, SQP can require several seconds to compute optimal trajectories in densely cluttered environments. Researchers in the past have exploited the underlying numerical structures to push the boundary of non-linear trajectory optimization \cite{toussaint_GN}. We hypothesize that we also need to explore if seemingly generic non-linear problems have some hidden convex structures to make further improvement. Our proposed optimizer makes essential contributions in this direction that can be summarized as follows




\noindent \textbf{Algorithmic:} For the first time, we present a convex approximation of the 3D trajectory optimization for FWVs. Our optimizer rests on the following novel ideas. First, we relax the collision avoidance and kinematic constraints as $l_2$  penalties to show that the resulting problem is multi-convex in three blocks of variables, namely position, forward velocity/acceleration, and sines and cosines of the orientation variables (See \cite{aks_iros20}, Def.2 for a quick primer on multi-convexity).  Second, we use Alternating Minimization (AM) technique to optimize over these different blocks of variables in a sequence. We show that each minimization is a quadratic program (QP) and moreover, some of the QPs even admit a symbolic solution. We augment some of the minimization steps with a Lagrange multiplier to drive the residuals of kinematic and collision avoidance constraints to zero. The multipliers are updated through Bregman iteration rule \cite{split_bergman}, \cite{admm_neural}. Finally, we use techniques from Alternating Direction Method of Multipliers (ADMM) \cite{admm_qp} to model the heading-angle rate constraints as convex quadratic penalties and reduce the constrained QP over heading angle to an unconstrained form. This further improves the computational efficiency of our optimizer. Refer Section \ref{compare_prior} for a summary of contribution over authors' prior works.

\noindent \textbf{Open-Source Implementation:} We provide an open-source implementation of our optimizer for review and to promote further research (\url{https://tinyurl.com/23vaj59m}). 

\noindent \textbf{State of the Art Performance:} We benchmark our optimizer against SQP implemented in ACADO \cite{acado}, the open-source library extensively used for rapid prototyping of trajectory optimization problems. Our optimizer shows a speed-up of $2-10$ times over ACADO's computation time while achieving solutions with $2-13$ times reduced control cost and similar residuals in the final positions.

\section{Preliminaries and Related Works}


\subsection{Symbols and Notations}
\noindent Small-case normal and bold font letters will represent scalars and vectors respectively. Bold-font upper-case letters will represent vectors. Table \ref{symbols} provides a list of important symbols used throughout the paper. We also define some symbols in the first place of their use. The superscript $T$ will be used to denote the transpose of matrices/vectors. The left superscript $k$ will be used to trace specific optimization variables over iterations in our optimizer. For example, ${^k}(.)$ represents the value of $(.)$ at iteration $k$. Planning horizon/steps and number of obstacles will be denoted by $n$ and $m$ respectively.

\subsection{Trajectory Optimization}
\noindent A standard 3D trajectory optimization for FWVs has the following form

\small
\begin{subequations}
\begin{align}
    \min \sum_t{\dot{v}(t)^2+\dot{\gamma}(t)^2+\dot{\phi}(t)^2}+ \left\Vert \begin{bmatrix}
    x(t_n)-x_f\\
    y(t_n)-y_f\\
    z(t_n)-z_f\\\end{bmatrix} \right\Vert_2^2 \label{cost_standard} \\
    \textbf{f}_{nh} = \textbf{0} \label{nonhol_standard} \\
    \dot{\psi}(t) = \frac{g}{v(t)}\tan\phi(t), \forall t, \label{psidot_nonhol_standard}\\
    (x(t), y(t), z(t))\in \mathcal{C}_{b} \label{boundary}\\
    v_{min} \leq v(t) \leq v_{max}, \vert\phi(t)\vert\leq \phi_{max}, \vert\gamma(t)\vert\leq \gamma_{max} \label{control_standard}\\
    -\frac{(x(t)-\xi_{xi})^2}{a_i^2}-\frac{(y(t)-\xi_{yi})^2}{b_i^2}-\frac{(z(t)-\xi_{zi})^2}{c_i^2}+1\leq 0, \label{coll_nonhol_standar}
\end{align}
\end{subequations}
\normalsize

\small
\begin{align}
   \textbf{f}_{nh} = \left \{ \begin{array}{lcr}
     \dot{x}(t) -v(t)\cos\psi(t)\cos\gamma(t),\\ 
 \dot{y}(t) -v(t)\sin\psi(t)\cos\gamma(t), \\
\dot{z}(t) +v(t)\sin\gamma(t),
\end{array} \right \}
\end{align}
\normalsize

\noindent The optimization variables are $(x(t), y(t), z(t), \psi(t), \gamma(t), \phi(t), v(t))$. Please see Table \ref{symbols} to note the physical attributes of these variables.  The FWV is supposed to be driven by forward acceleration ($\dot{v}(t)$), bank ($\dot{\phi}(t)$) and flight-path angle ($\dot{\gamma}(t)$) rate. The cost function (\ref{cost_standard}) is a combination of sum of squares of the control inputs at different time instants and the goal-reaching cost with respect to the final position ($x_f, y_f, z_f$). Equality constraints (\ref{nonhol_standard})-(\ref{psidot_nonhol_standard}) stem from the so-called non-holonomic kinematic model of FWVs  with bank-to-turn assumption \cite{randal_implement}. Constraint (\ref{boundary}) enforces the initial boundary condition on position and its derivatives. Inequality (\ref{control_standard}) represent the bounds on the forward velocity, the bank angle $\phi(t)$ and the flight-path angle $\gamma(t)$. To maintain flight, $v_{min}$ should be strictly greater than a specified positive value. The set of inequalities (\ref{coll_nonhol_standar}) are the collision avoidance constraints with respect to the $i^{th}$ obstacle with position $(\xi_{xi}, \xi_{yi}, \xi_{zi})$. Herein, we have made the assumption that the FWV is a sphere while the obstacles are axis-aligned ellipsoids. The constants $a_i, b_i, c_i$ are the dimensions of the ellipsoids inflated by the radius of the FWV.

\begin{table}[!t]
\scriptsize
\centering
\caption{Important Symbols  } \label{symbols}
\begin{tabular}{|p{3.9cm}|p{4cm}|p{5cm}|p{5cm}|}\hline
$x(t), y(t), z(t), \psi(t), \phi(t), \gamma(t)$ & Position, heading, bank (roll) and flight-path angle (pitch) respectively of FWV at time $t$. See accompanying video\\\hline
\mbox{$\xi_{xi}, \xi_{yi}, \xi_{zi}$}& Position of the $i^{th} $obstacle\\ \hline
\mbox{$\alpha_i(t), \beta_i(t), d_i(t)$ }& Parameters associated with our collision avoidance model. Refer text for details.\\ \hline
\end{tabular}
\vspace{-0.6cm}
\normalsize
\end{table}

\noindent Optimization (\ref{cost_standard})-(\ref{coll_nonhol_standar}) involves a search in the space of functions. A more tractable form can be achieved by assuming some parametric representation for these functions such as polynomials (e.g see (\ref{param})). Under, this reduction, (\ref{cost_standard})-(\ref{coll_nonhol_standar}) becomes a standard non-linear programming problem over the coefficients of the polynomial.

\subsection{Existing Approaches}
\noindent \textbf{Single Vs Multiple Shooting} There are two broad class of approaches for solving an optimal control problem of the form (\ref{cost_standard})-(\ref{coll_nonhol_standar}), namely single and multiple shooting. The single shooting approaches will assume a parametrization for the control inputs $\dot{v}(t), \dot{\phi}(t), \dot{\gamma}(t)$ and represent the states as analytical functions of the control by integrating (\ref{nonhol_standard})-(\ref{psidot_nonhol_standard}). As a result, the state variables are eliminated from the optimization process. In contrast, multiple shooting retains both state and control inputs as independent variables tied together by the kinematic constraints. Multiple shooting leads to a sparse structure in the optimization problem and has become the state of art for FWVs \cite{multiple_shoot_fwv_1}, \cite{multiple_shoot_fwv_2}.

\noindent \textbf{Critical Gaps } Most works on FWV trajectory optimization target applications like path tracking/following/landing in obstacle free space \cite{multiple_shoot_fwv_1}, \cite{multiple_shoot_fwv_2}, \cite{fwv_landing}. Collision Avoidance is often achieved by sampling based planners augmented with smoothing operations to satisfy kinematic constraints \cite{bry_ijrr}, \cite{fwv_gd}. Alternately, collision avoidance is achieved through reactive planning \cite{fwv_sense_avoid}. Only recently, works like \cite{sqp_fwv_adhikari}, \cite{gusto} have incorporated collision avoidance within trajectory optimization of FWVs. However, \cite{sqp_fwv_adhikari} only considers sparsely filled environments with $2-4$ obstacles in their experiments.

Our optimizer adopts the multiple-shooting paradigm. However, in sharp contrast to existing works, our optimizer never linearizes the underlying non-linear constraints. As a result, it works with arbitrary initialization on all our considered benchmarks and maintains near real-time performance in densely cluttered environments.

\subsection{Comparisons with Author's Prior Work} \label{compare_prior}
\noindent Our optimizer builds on the preliminary work \cite{aks_icra20} that dealt with the 2D trajectory optimization of FWVs. The 2D case has a simpler bi-convex structure compared to the multi-convex structure present in the 3D problem. We use the collision avoidance model proposed in \cite{aks_iros20} for affine differentially flat systems. Thus, our optimizer validates the efficacy of these models for non-linear systems such as FWVs. Our optimizer also differs from our cited prior works in the manner we use the Lagrange multiplier to drive the constraint residuals to zero. We defer further discussion into this till Section \ref{lagrange_section}


\subsection{Trajectory Parameterization}
\noindent Our optimizer parameterizes the position and heading trajectory in the following manner

\begin{equation}
\begin{bmatrix}
x(t_1)\\
x(t_2)\\
\dots\\
x(t_n)
\end{bmatrix} = \textbf{P}\textbf{c}_{x}, \begin{bmatrix}
\psi(t_1)\\
\psi(t_2)\\
\dots\\
\psi(t_n)
\end{bmatrix} = \textbf{P}\textbf{c}_{\psi},
\label{param}
\end{equation}

\noindent where, $\textbf{P}$ is a matrix formed with time-dependent basis functions (e.g polynomials) and $\textbf{c}_{x}, \textbf{c}_{\psi}$ are the coefficients associated with the basis functions. Similar expressions can be written for $y(t), z(t)$ as well. We can also express the derivatives in terms of $\dot{\textbf{P}}, \ddot{\textbf{P}}$.

\section{Main Results}


\subsection{Building Blocks}
\noindent \textbf{Collision Avoidance Model:}
\noindent Our optimizer takes a departure from the conventional quadratic collision avoidance model (\ref{coll_nonhol_standar}) and instead adopts the form proposed in our prior work \cite{aks_iros20}.  Herein collision avoidance is modeled through non-linear equality constraints of the form $\textbf{f}_c = \textbf{0}$, where

\begin{align}
    \textbf{f}_{c} = \left \{ \begin{array}{lcr}
x(t) -\xi_{xi}-a_id_{i}(t)\sin\beta_{i}(t)\cos\alpha_{i}(t) \\
y(t) -\xi_{yi}-b_id_{i}(t)\sin\beta_{i}(t)\sin\alpha_{i}(t)\\ 
z(t) -\xi_{zi}-c_id_{i}(t)\cos\beta_{i}(t) \\
\end{array} \right \}
\label{sphere_proposed}
\end{align}
\normalsize

\noindent As can be noted, $\textbf{f}_c$ is a polar representation of the euclidean distance between the robot and the obstacle. The variables $\alpha_i(t), \beta_i(t)$ are the 3D solid angle of the line of sight vector connecting the robot and the $i^{th}$ obstacle. The length of this vector is given by $d_i(t)\sqrt{a_i^2+b_i^2+c_i^2}$. Thus, collision avoidance can be enforced by ensuring $d_{i}(t)\geq 1$. It should be noted that in (\ref{sphere_proposed}), $\alpha_i(t), \beta_i(t), d_i(t)$ are unknown variables that are obtained by our optimizer along with other trajectory variables.

There are several advantages of using (\ref{sphere_proposed}) as the collision avoidance model over (\ref{coll_nonhol_standar}). First, the equality constraints $\textbf{f}_c = \textbf{0}$ has a natural relaxation in the form of $l_2$ penalty, which we show later can be efficiently optimized using the AM technique. Second, as the number of obstacles and planning steps increase, the computational cost associated with the $l_2$ penalty increases at a much lower rate than that induced by the rise in the number of constraints (\ref{coll_nonhol_standar}) \cite{aks_iros20}.

\noindent \textbf{Change in Smoothness Cost and Eliminating Bank-angle}

\noindent Our smoothness cost differs from the standard choice presented in (\ref{cost_standard}). We penalize the squared acceleration components $(\ddot{x}^2(t), \ddot{y}^2(t), \ddot{z}^2(t), \ddot{\psi}^2(t)  )$ instead of $(\dot{v}_t^2, \dot{\phi}(t)^2, \dot{\gamma}(t)^2)$. As shown latter, this leads to key computational gains as optimizations over $v(t), \gamma(t)$ reduces to just evaluating closed form symbolic formulae. Furthermore, we validate empirically in Section \ref{sim} that our chosen representation is still rich enough to ensure smooth evolution of forward acceleration, flight-path and banking angle.

We also note that the banking angle $\phi(t)$, its bound constraints, and (\ref{psidot_nonhol_standard}) can be eliminated by introducing an additional constraint of the form $\vert \dot{\psi}(t)\vert \leq \frac{g}{v}\tan\phi_{max}$. This will ensure that the FWV always attain the heading angle rate compatible with the banking limit. The bank-angle and its derivatives can be obtained post optimization through heading-angle rate and forward velocity (see (\ref{psidot_nonhol_standard})). This reduction is not surprising as FWVs are deferentially flat and thus we can recover control inputs from state trajectory derivatives.

\subsection{Reformulation and Relaxation}

\noindent We are now in a position to start developing our main theoretical results. We begin by presenting the following reformulation of (\ref{cost_standard})-(\ref{coll_nonhol_standar}) based on the discussions of the previous two sections.
\small
\begin{subequations}
\begin{align}
    \min \sum_t \overbrace{\ddot{x}(t)^2+\ddot{y}(t)^2+\ddot{z}(t)^2 + \ddot{\psi}(t)^2+\left \Vert\begin{bmatrix}
    x(t_n)-x_f\\
    y(t_n)-y_f\\
    z(t_n)-z_f\\\end{bmatrix} \right\Vert_2^2}^{c(x(t), y(t), z(t), \psi(t))} \label{cost_proposed}\\
    \textbf{f}_{nh} = \textbf{0} \label{nonhol_proposed}\\
     v_{min} \leq v(t) \leq v_{max}, \vert \gamma(t) \vert \leq \gamma_{max} \label{bounds_proposed} \\
     \vert \dot{\psi}(t)\vert \leq \frac{g}{v(t)}\tan\phi_{max} \label{tunrate_proposed}\\
     \textbf{f}_c = \textbf{0} \label{coll_proposed}\\
     d_i(t)\geq 1, \beta_{i}(t) \in [0, \pi],  \alpha_{i}(t) \in [-\pi, \pi] \label{coll_slack_proposed}
\end{align}
\end{subequations}
\normalsize

The variables in our reformulated problem consists of FWV trajectory  $(x(t), y(t), v(t), \psi(t), \gamma(t))$ and the obstacle avoidance parameters $(\alpha_i(t), \beta_i(t), d_i(t))$. Our optimizer solves (\ref{cost_proposed})-(\ref{coll_proposed}) by constructing the following augmented Lagrangian, wherein the kinematic and collision-avoidance constraints are relaxed as quadratic penalties.

\small
\begin{align}
    \mathcal{L} = \sum_t c(x(t), y(t), z(t), \psi(t)) +\frac{\rho_{nh}}{2}\left\Vert\textbf{f}_{nh}\right\Vert_2^2+\frac{\rho_c}{2}\left\Vert \textbf{f}_c\right\Vert_2^2 \label{aug_lagrange}
\end{align}
\normalsize

\newtheorem{remark}{Remark}
\begin{remark}\label{remark_multiconvexity}
For a given $\psi(t), \gamma(t), v(t)$, the Lagrangian (\ref{aug_lagrange}) is convex 
in each of $x(t), y(t), z(t)$ and their derivatives.
\end{remark}

\begin{remark}\label{remark_surrogate}
For a given $(x(t), y(t), z(t))$, $\gamma(t)$, and $v(t)$, the non-convex penalty $\left\Vert\textbf{f}_{nh}\right\Vert_2^2$ is convex in the space of $ (\cos\psi(t), \sin\psi(t))$ \cite{aks_icra20}. Similarly, for a fixed $\psi(t)$, $\left\Vert\textbf{f}_{nh}\right\Vert_2^2$ is convex in the space of $(\cos\gamma(t), \sin\gamma(t))$.
\end{remark}

\noindent The computational structures highlighted in Remarks \ref{remark_multiconvexity} and \ref{remark_surrogate} forms the foundation of our optimizer. At the same time, they also give a glimpse of why an AM technique will be appropriate for optimizing over the Lagrangian (\ref{aug_lagrange}). Specifically, we can reduce each minimization step in AM to a convex problem by choosing the optimization blocks carefully. We delve into further details in the following subsection.


\subsection{AM Based Solution}
\noindent Algorithm \ref{algo_1} presents the various steps in AM-based minimization of (\ref{aug_lagrange}) subject to constraints (\ref{bounds_proposed}), (\ref{tunrate_proposed}) and (\ref{coll_slack_proposed}). Please recall that $k$ represents the iteration index, and when used as a left superscript, it tracks the value of the respective variable over iterations. The Algorithm starts with initialization of ${^k}d_{i}(t), {^k}\alpha_{i}(t), {^k}\beta_{i}, {^k\psi(t), {^k}\gamma(t), {^k}v(t)} $ for the first iteration. At each of steps (\ref{step_x_3d})-(\ref{step_d}), the minimization is done over only one variable while the others are held fixed at values obtained at the previous iteration or at the earlier steps of the same iteration. 

\subsubsection{Step (\ref{step_x_3d})} This specific optimization can be framed as the following convex QP

\small
\begin{subequations}
\begin{align}
    \min \frac{1}{2}\textbf{c}_x^T(\textbf{Q}_x+\rho_c\textbf{A}_{f_c}^T\textbf{A}_{f_c}+\rho_{nh}\dot{\textbf{P}}^T\dot{\textbf{P}})\textbf{c}_x\nonumber \\
    +(\textbf{q}_x+{^k}\textbf{b}_{f_c}^x+{^k}\textbf{b}_{nh}^x)^T\textbf{c}_x,\\
    \textbf{A}_{eq} \textbf{c}_x = \textbf{b}_{eq}^x.
    \label{step_x_QP}
\end{align}
\end{subequations}
\normalsize

\noindent The matrices can be derived in the following manner using the trajectory parametrizations defined in (\ref{param}). In the following $\textbf{P}_i, \dot{\textbf{P}}_i, \ddot{\textbf{P}}_i$ represent the $i^{th}$ row of the respective matrices.

\small
\begin{align}
x(t)\in \mathcal{C}_{b}  = \overbrace{\begin{bmatrix}
\textbf{P}_0|
\dot{\textbf{P}}_0|
\ddot{\textbf{P}}_0
\end{bmatrix}}^{\textbf{A}_{eq}} = \overbrace{\begin{bmatrix}
x(t_0)|
\dot{x}(t_0)|
\ddot{x}(t_0)
\end{bmatrix}}^{\textbf{b}_{eq}^x}
\label{a_eq_mat_x}
\end{align}
\normalsize

\small
\begin{align}
\sum_t \ddot{x}(t)^2+(x(t_n)-x_f)^2 \nonumber \\ \Rightarrow  \frac{1}{2}\textbf{c}_x\overbrace{(\ddot{\textbf{P}}^T \ddot{\textbf{P}}+\textbf{P}_n^T\textbf{P}_n)}^{\textbf{Q}_x}\textbf{c}_x
+\overbrace{(-x_f\textbf{P}_n^T}^{\textbf{q}_x})^T\textbf{c}_x
\label{ste_x_cost}
\end{align}
\normalsize

\small
\begin{align}
    \sum_t\frac{\rho_{nh}}{2} (\dot{x}(t)-{^k}v(t)\cos{^k}\psi(t)\cos{^k}\gamma(t))^2 \Rightarrow \frac{1}{2}\textbf{c}_x^T(\rho_{nh}\dot{\textbf{P}}^T\dot{\textbf{P}})\textbf{c}_x\nonumber \\
    +(\overbrace{-\rho_{nh}\dot{\textbf{P}}^T({^k}\textbf{v}\cos{^k}\boldsymbol{\psi}\cos{^k}\boldsymbol{\gamma})}^{{^k}\textbf{b}_{nh}^x})^T\textbf{c}_x,
    \label{step_x_nonhol}
\end{align}
\normalsize

\small 
\begin{align}
    \sum_{t,i}\frac{\rho_{c}}{2}(x(t) -\xi_{xi}(t)-a_i{^k}d_{i}(t)\sin{^k}\beta_{i}(t)\cos{^k}\alpha_{i}(t))^2 \nonumber \\
    \Rightarrow \frac{1}{2} \textbf{c}_x^T(\rho_c\textbf{A}_{f_c}^T\textbf{A}_{f_c})\textbf{c}_x+(\overbrace{-\rho_c\textbf{A}_{f_c}^T(\boldsymbol{\xi}_{x}+\textbf{a}{^k}\textbf{d}\cos{^k}\boldsymbol{\alpha}\sin{^k}\boldsymbol{\beta}  ) }^{{^k}\textbf{b}_{f_c}^x})^T\textbf{c}_x
    \label{step_x_coll}
\end{align}
\normalsize


\noindent In (\ref{step_x_nonhol}), ${^k}\textbf{v}$ is formed by stacking ${^k}v(t)$ at different time instants. Similar process follows for ${^k}\boldsymbol{\psi}, {^k}\boldsymbol{\gamma}, {^k}\textbf{d}, {^k}\boldsymbol{\alpha}, {^k}\boldsymbol{\beta}$. The $\boldsymbol{\xi}_{x}$ is formed by stacking the $x$ position of all the obstacles. The matrix $\textbf{A}_{f_c}$ is formed by stacking $\textbf{P}$ vertically as many times as the number of obstacles.

The equality constrained QP (\ref{step_x_QP}) reduces to solving the following linear equations, wherein $\boldsymbol{\mu}_x$ represents the dual variables associated with the equality constraints. The l.h.s of (\ref{kkt_step_x}) is called the Karush-Kuhn Tucker (KKT) Matrix.

\scriptsize
\begin{align}
    \begin{bmatrix}
    (\textbf{Q}_x+\rho_c\textbf{A}_{f_c}^T\textbf{A}_{f_c}+\rho_{nh}\dot{\textbf{P}}^T\dot{\textbf{P}}  ) & \textbf{A}_{eq}^T\\
    \textbf{A}_{eq} & \textbf{0}
    \end{bmatrix} \begin{bmatrix}
    \textbf{c}_x\\
    \boldsymbol{\mu}_x
    \end{bmatrix} = \begin{bmatrix}
    -\textbf{q}_x-{^k}\textbf{b}_{f_c}^x-{^k}\textbf{b}_{nh}^x\\
    \textbf{b}_{eq}^x
    \end{bmatrix}
\label{kkt_step_x}
\end{align}
\normalsize

\noindent Similar steps can be derived for minimization over $y(t), z(t)$ on steps (\ref{step_y_3d})-(\ref{step_z_3d}).

\begin{remark} \label{kkt_remark}
During (\ref{step_y_3d})-(\ref{step_z_3d}), we obtain the same KKT matrix as (\ref{kkt_step_x}). Thus, we need to compute its factorization only once at (\ref{step_x_3d}) and use the same on steps (\ref{step_y_3d})-(\ref{step_z_3d}).
\end{remark}

\noindent The comments in Remark \ref{kkt_remark} follows from the fact $x(t), y(t), z(t)$ are parameterized by the same basis functions $\textbf{P}$.

\subsubsection{Step (\ref{step_psi})} The first line in (\ref{step_psi}) is non-convex. Specifically, the non-convexity stems from the first two terms that involve $\cos\psi(t), \sin\psi(t)$. But using the insights from Remark \ref{remark_surrogate}, we can replace these terms with a convex surrogate proposed in our earlier work \cite{aks_icra20}, \cite{aks_ecc20}. As a result, minimization (\ref{step_psi}) also reduces to a QP with the following form.

\vspace{-0.4cm}
\small
\begin{subequations}
\begin{align}
    \min \frac{1}{2}\textbf{c}_{\psi}^T\textbf{Q}_{\psi}\textbf{c}_{\psi}
    +{^k}\textbf{q}_{\psi}^T\textbf{c}_{\psi}, \label{psi_QP_cost} \\
    \textbf{A}_{in}\textbf{c}_{\psi}\leq {^k}\textbf{b}_{in}
    \label{psi_QP_con}
\end{align}
\end{subequations}
\normalsize
\vspace{-0.5cm}
\small
\begin{align}
    \textbf{A}_{in} = \begin{bmatrix}
    \dot{\textbf{P}}\\
    -\dot{\textbf{P}}
    \end{bmatrix}, {^k}\textbf{b}_{in} = \begin{bmatrix}
    \frac{g}{{^k}\textbf{v}}\tan\phi_{max}\\
    \frac{g}{{^k}\textbf{v}}\tan\phi_{max}
    \end{bmatrix} 
\end{align}
\normalsize

\small
\begin{align}
    \sum_t \frac{\rho_{nh}}{2}\Vert \arctan2({^{k+1}}\dot{y}(t), {^{k+1}}\dot{x}(t))-\psi(t)\Vert_2^2+\ddot{\psi}^2(t) 
    \Rightarrow   \nonumber \\\frac{1}{2} \textbf{c}_{\psi}^T \overbrace{(\ddot{\textbf{P}}^T \ddot{\textbf{P}}+\rho_{nh}\textbf{P}^T\textbf{P})}^{\textbf{Q}_{\psi}}\textbf{c}_{\psi}+(\overbrace{-\rho_{nh}\textbf{P}^T \arctan2({^{k+1}}\dot{\textbf{y}}, {^{k+1}}\dot{\textbf{x}})}^{{^k}\textbf{q}_{\psi}})^T\textbf{c}_{\psi},
    \label{step_psi_cost}
\end{align}
\normalsize

\noindent where, as before, ${^{k+1}}\dot{\textbf{x}}, {^{k+1}}\dot{\textbf{y}}$ is formed by stacking values at different time instants.

\subsubsection{Step (\ref{step_gama})} The key idea here is that for a given ${^{k+1}}\dot{x}(t), {^{k+1}}\dot{y}(t), {^{k+1}}\dot{z}(t), {^{k+1}}\psi(t), {^{k+1}}v(t)$, $\gamma(t)$ at various time instants can be treated as decoupled from each other. This reduces (\ref{step_gama}) to $n$ de-coupled single variable QP in terms of $(\cos\gamma(t), \sin\gamma(t))$ (recall Remark \ref{remark_surrogate}). Thus, differentiating each of these problems with respect to $\gamma(t)$ leads to an affine expression in term of $(\cos\gamma(t), \sin\gamma(t))$. This when equated to zero leads to a closed form symbolic solution. The bounds on $\gamma(t)$ are ensured by simple clipping.

\subsubsection{Step (\ref{step_v})} For a given ${^{k+1}}\dot{x}(t), {^{k+1}}\dot{y}(t), {^{k+1}}\dot{z}(t) $, computing the forward velocity is trivial and the formula is presented in (\ref{step_v}). We clip the obtained values to satisfy the bounds and heading-angle rate constraint.

\subsubsection{Step (\ref{step_alpha}) } Similar to previous step, $\alpha_i(t), \beta_i(t)$ at various time instants are decoupled from each other. Thus, minimization (\ref{step_alpha}) decomposes to $n$ parallel problems, each of which pertains to projecting ${^{k+1}}x(t), {^{k+1}}y(t), {^{k+1}}z(t)$ onto a sphere with center $(\xi_{xi}, \xi_{yi}, \xi_{zi})$. As shown, the solution can be expressed as a formula. The constraints on $\alpha(t)$ are satisfied by construction because of $\arctan2(.)$ used in (\ref{step_alpha}). Bounds on $\beta(t)$ is ensured by clipping.

\subsubsection{Step (\ref{step_d})} This step has the same parallel structure as (\ref{step_alpha}) and thus leads to $n$ parallel single-variable QP. The constraints on $d_i(t)$ are enforced by a simple clipping in the interval $[1, \infty)$.

\subsection{Incorporating Lagrange Multipliers } \label{lagrange_section}
\noindent Algorithm \ref{algo_1} has one critical limitation. As the iteration progress, the residuals of $\Vert \textbf{f}_{nh}\Vert_2^2 $,  $\Vert \textbf{f}_{c}\Vert_2^2 $ decrease but driving them arbitrarily close to zero, in theory requires infinitely large $\rho_{nh}$, $\rho_{c}$. This is one of the quintessential limitation of modeling equality constraints as $l_2$ penalties and forms the motivation behind the augmented Lagrangian approaches. To counter this, we introduce the so-called Lagrange multiplier $\boldsymbol{\lambda}_x$ and modify the cost function (\ref{step_x_QP}) in the following manner.

\small
\begin{align}
\frac{1}{2}\textbf{c}_x^T(\textbf{Q}_x+\rho_c\textbf{A}_{f_c}^T\textbf{A}_{f_c}+\rho_{nh}\dot{\textbf{P}}^T\dot{\textbf{P}})\textbf{c}_x\nonumber \\
    +(\textbf{q}_x+{^k}\textbf{b}_{f_c}^x+{^k}\textbf{b}_{nh}^x-{^k}\boldsymbol{\lambda}_x)^T\textbf{c}_x.
\end{align}
\normalsize

\noindent The multiplier is updated at the end of each iteration based on the residuals of the $l_2$ penalties by the following rule.

\small
\begin{subequations}
\begin{align}
{^{k+1}}\boldsymbol{\lambda}_x = {^k}\boldsymbol{\lambda}_x-\rho_{nh}\dot{\textbf{P}}^T{^{k+1}}\textbf{r}_{nh}^x-\rho_c\textbf{A}_{f_c}^T{^{k+1}}\textbf{r}_{f_c}^x\\
    {^{k+1}}\textbf{r}_{nh}^x = \dot{\textbf{P}} {^{k+1}}\textbf{c}_x -{^{k+1}}\textbf{v}\cos{^{k+1}}\boldsymbol{\psi}\cos{^{k+1}}\boldsymbol{\gamma},\\
    {^{k+1}}\textbf{r}_{f_c}^x = (\dot{\textbf{A}_{f_c}} {^{k+1}}\textbf{c}_x -\boldsymbol{\xi}_x-\textbf{a}{^{k+1}}\textbf{d}\cos{^{k+1}}\boldsymbol{\alpha}\sin{^{k+1}}\boldsymbol{\beta},
\end{align}
\end{subequations}
\normalsize

\noindent Following a similar process as above, multipliers are also introduced for (\ref{step_y_3d})-(\ref{step_psi}).

The Lagrange multiplier usage follows from Bregman iteration (BI) \cite{split_bergman}, \cite{admm_neural} and is strikingly different from our prior work  \cite{aks_iros20}, \cite{aks_icra20} that followed the closely related ADMM based process. In ADMM, a multiplier needs to be introduced for each element of $\textbf{f}_{nh}, \textbf{f}_c$ at each time instant. Thus, we need a Lagrange multiplier of dimension $3n$ for each of the constraints. In contrast, BI needs only a single multiplier whose dimension is equal to the dimension of $\textbf{c}_x$. The second motivation for shifting to BI is that our experiments have shown that it converges faster than our prior ADMM based work. Similar observations were also reported in \cite{admm_neural} that also applied BI to a highly non-convex problem.

\subsection{Unconstrained Variant}

\noindent Step (\ref{step_psi}) requires us to solve a constrained QP at each iteration. This can be computationally costly given that AM and Lagrange multipliers based optimizers such as Algorithm \ref{algo_1} require tens of iterations to converge to a low residual solution \cite{boyd_admm}. Thus, it is important to make each iteration as inexpensive as possible. To this end, we use the ADMM based technique proposed in \cite{admm_qp} and incorporate the heading-angle rate constraints as quadratic costs. This involves modifying the QP (\ref{psi_QP_cost})-(\ref{psi_QP_con}) into the unconstrained form (\ref{psi_uncon}), where $\boldsymbol{\lambda}_{in}$ is the Lagrange multiplier and $\textbf{s}_{in}$ are slack variables used to convert inequality constraints into an equality form. The multipliers and slack variables are updated based on constraint residuals, as shown in (\ref{s_update})-(\ref{lamda_in_update}).

\begin{algorithm*}[!h]
\centering
 \caption{Alternating Minimization based 3D Trajectory Optimization for FWVs }\label{algo_1}
    \begin{algorithmic}[1]   
\State Initialize   ${^k}d_{i}(t), {^k}\alpha_{i}(t), {^k}\beta_{i}(t), {^k\psi(t), {^k}\gamma(t), {^k}v(t)} $ at $k = 0$
\While {$k\leq maxiter$ \text{ or till norm of the residuals are below some threshold}}
\begin{subequations}
\small
\begin{align}
{^{k+1}}x(t) = \arg\min_{x(t) \in \mathcal{C}_{b}} \sum_{t}\ddot{x}(t)^2+\frac{\rho_{nh}}{2} (\dot{x}(t)-{^k}v(t)\cos{^k}\psi(t)\cos{^k}\gamma(t))^2+\sum_{t,i}\frac{\rho_{c}}{2}(x(t) -\xi_{xi}-a_i{^k}d_{i}(t)\sin{^k}\beta_{i}(t)\cos{^k}\alpha_{i}(t))^2,  \label{step_x_3d}\\ 
{^{k+1}}y(t) = \arg\min_{y(t)\in \mathcal{C}_b} \sum_{t}\ddot{y}(t)^2+\frac{\rho_{nh}}{2} (\dot{y}(t)-{^k}v(t)\sin{^k}\psi(t)\cos{^k}\gamma(t))^2+\sum_{t,i}\frac{\rho_{c}}{2}(y(t) -\xi_{yi}-b_i{^k}d_{i}(t)\sin{^k}\beta_{i}(t)\sin{^k}\alpha_{i}(t))^2,  \label{step_y_3d} \\
{^{k+1}}z(t) = \arg\min_{z(t) \in \mathcal{C}_b} \sum_{t}\ddot{z}(t)^2+\frac{\rho_{nh}}{2} (\dot{z}(t)+{^k}v(t)\sin{^k}\gamma(t))^2+\sum_{t,i}\frac{\rho_{c}}{2}(z(t) -\xi_{zi}-c_i{^k}d_{i}(t)\cos{^k}\beta_{i}(t))^2, \label{step_z_3d}\\
 {^{k+1}}\psi(t) = \arg\min \sum_t\frac{\rho_{nh}}{2} ({^{k+1}}\dot{x}(t)-{^k}v(t)\cos\psi(t)\cos{^k}\gamma(t))^2+\frac{\rho_{nh}}{2} ({^{k+1}}\dot{y}(t)-{^k}v(t)\sin\psi(t)\cos{^k}\gamma(t))^2+\ddot{\psi}(t)^2, \nonumber \text{subject to} \\ \vert \dot{\psi}(t)\vert \leq \frac{g}{{^{k+1}}v(t)}\tan\phi_{max}  \nonumber  \\
 \Rightarrow {^{k+1}}\psi(t) = \arg \min \sum_t \frac{\rho_{nh}}{2}\Vert \arctan2({^{k+1}}\dot{y}(t), {^{k+1}}\dot{x}(t))-\psi(t)\Vert_2^2+\ddot{\psi}^2(t), \vert \dot{\psi}(t)\vert \leq \frac{g}{{^{k+1}}v(t)}\tan\phi_{max} \label{step_psi}  \\
 {^{k+1}}\gamma(t) = \arg\min_{\gamma(t)} \frac{\rho_{nh}}{2}\sum_t ({^{k+1}}\dot{x}(t)-{^k}v(t)\cos{^{k+1}}\psi(t)\cos\gamma(t))^2+ ({^{k+1}}\dot{y}(t)-{^k}v(t)\sin{^{k+1}}\psi(t)\cos\gamma(t))^2\nonumber\\+ ({^{k+1}}\dot{z}(t)+{^k}v(t)\sin\gamma(t))^2 \nonumber \\
 \Rightarrow {^{k+1}}\gamma(t) = \arctan2(-{^{k+1}}\dot{z}(t), {^{k+1}}\dot{x}(t)\cos{^{k+1}}\psi(t)+{^{k+1}}\dot{y}\sin{^{k+1}}\psi(t), \text{clipped to } (-\gamma_{max}, \gamma_{max}) 
\label{step_gama}\\
 {^{k+1}}v(t) = \sqrt{{^{k+1}}\dot{x}(t)^2+{^{k+1}}\dot{y}(t)^2+{^{k+1}}\dot{z}(t)^2}, \text{clipped to } (v_{min}, min(v_{max}, \frac{g\tan\phi_{max}}{\vert{^{k+1}}\dot{\psi}(t)\vert}   )
 \label{step_v}\\ 
 {^{k+1}}\alpha_{i}(t), {^{k+1}}\beta_{i}(t) = \arg\min_{\alpha_{i}, \beta_{i}} \sum_{i, t} \frac{\rho_c}{2}({^{k+1}}x(t) -\xi_{xi}-a_i{^k}d_{i}(t)\sin\beta_{i}(t)\cos\alpha_{i}(t))^2\nonumber 
 \\
 +\frac{\rho_c}{2}({^{k+1}}y(t) -\xi_{yi}-b_i{^k}d_{i}(t)\sin\beta_{i}(t)\sin\alpha_{i}(t))^2  
 +\frac{\rho_c}{2}({^{k+1}}z(t)-\xi_{zi}-c_i{^k}d_{i}(t)\cos\beta_{i}(t))^2 \color{black}\label{step_alpha}\\ 
 \Rightarrow   {^{k+1}}\alpha_{i}(t) = \arctan2({^{k+1}}y(t)-\xi_{yi}, {^{k+1}}x(t)-\xi_{xi}),
 {^{k+1}}\beta_{i}(t) = \arctan2(\frac{{^{k+1}}x(t)-\xi_{xi}}{a_i \cos{{^{k+1}}\alpha_{i}(t)}}, \frac{{^{k+1}}z(t)-\xi_{zi}}{c_i} ) \label{sphere_project} \\
 {^{k+1}}d_{i}(t) = \arg\min_{d_{i}} \sum_{i, t}\frac{\rho_c}{2}({^{k+1}}x(t) -\xi_{xi}-a_id_{i}(t)\sin{^{k+1}}\beta_{i}(t)\cos{^{k+1}}\alpha_{i}(t))^2 \nonumber \\
 +\frac{\rho_c}{2}({^{k+1}}y(t) -\xi_{yi}-b_id_{i}(t)\sin{^{k+1}}\beta_{i}(t)\sin{^{k+1}}\alpha_{i}(t)+)^2
 +\frac{\rho_c}{2}({^{k+1}}z(t) -\xi_{zi}-c_id_{i}(t)\cos{^{k+1}}\beta_{i}(t))^2, d_i(t)\geq 1 \label{step_d} 
\end{align}
\end{subequations}
\normalsize
\EndWhile
\end{algorithmic}
\end{algorithm*}

\small
\begin{align}
     \min \frac{1}{2}\textbf{c}_{\psi}^T\textbf{Q}_{\psi}\textbf{c}_{\psi}+{^k}\textbf{q}_{\psi}^T\textbf{c}_{\psi}+\frac{\rho_{in}}{2}\left\Vert \textbf{A}_{in}\textbf{c}_{\psi}-{^k}\textbf{b}_{in}+{^k}\textbf{s}_{in}+\frac{\boldsymbol{{^k}\lambda}}{\rho_{in}}\right\Vert_2^2
     \label{psi_uncon}
\end{align}

\begin{subequations}
\begin{align}
{^{k+1}}\textbf{s}_{in} = \max({\textbf{0}, -\textbf{A}_{in}{^{k+1}}\textbf{c}_{\psi}+{^{k+1}}\textbf{b}_{in}}-\frac{{^k}\boldsymbol{\lambda}_{in}}{\rho_{in}}) \label{s_update}  \\
{^{k+1}}\boldsymbol{\lambda}_{in} = {^k}\boldsymbol{\lambda}_{in}
+\rho_{in}(\textbf{A}_{in}{^{k+1}}\textbf{c}_{\psi}-{^{k+1}}\textbf{b}_{in}+{^{k+1}}\textbf{s}_{in}) \label{lamda_in_update}
\end{align}
\end{subequations}
\normalsize

\begin{figure}
     \centering
     \includegraphics[scale=0.4]{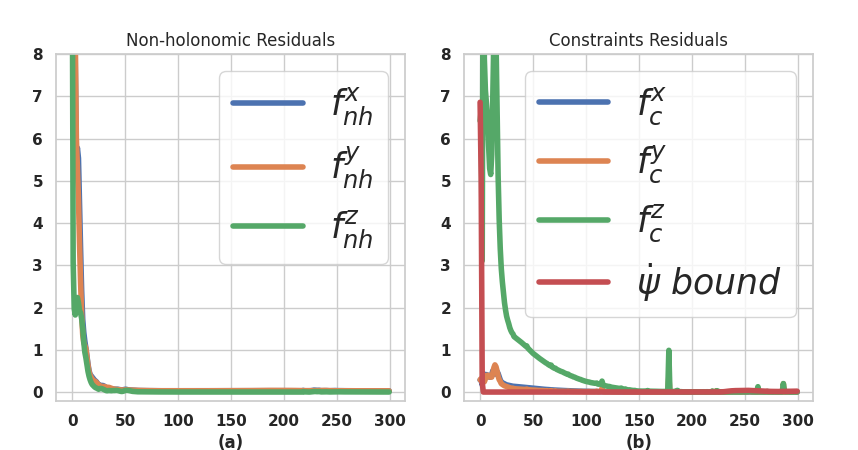}
    \caption{Mean residuals of kinematic, collision avoidance and heading-angle rate constraints observed across all the benchmarks. Note that constraints of $v(t), \gamma(t)$ are satisfied exactly by clipping to min/max iterval at each iteration.}
    \label{convergence}
    \vspace{-0.5cm}
\end{figure}

\subsection{Traversal Time Heuristic}
\noindent As mentioned earlier, FWVs need to maintain a minimum forward velocity to maintain flight. Furthermore, due to aerodynamic reasons, the difference between $v_{min}$ and $v_{max}$ is kept small. Due to these reasons, the choice of traversal time becomes very critical. For example, choosing an arbitrary large value might be favorable from the actuation standpoint, but would lead to longer arc-length trajectory, wherein the FWV will do loops around the goal point. We provide a simple heuristic solution to this end. We run a few iterations of Algorithm \ref{algo_1} with just collision avoidance constraints and the smoothness cost and ignoring the kinematic constraints and bounds. We compute the arc-length of the resulting trajectory and divide it by $v_{min}$ to get an approximation of planning time.

\section{Simulation and Bench-marking}\label{sim}
\noindent \textbf{Implementation Details:} We prototyped Algorithm \ref{algo_1} in C++ using Eigen as our linear algebra library. Our test benchmark is an urban environment with 13 buildings acting as obstacles (see accompanying video). We create 20 different configurations by varying start/goal positions and initial heading angle. For each configuration, we additionally create two different variants with 50 and 100 planning steps.

\noindent \textbf{Convergence Validation: } Fig.\ref{convergence} shows the mean residuals of various constraints observed across iterations over all the considered benchmarks. As shown, residuals decrease over iterations and on average, 300 iterations always proved enough to obtain residuals in the order of $10^{-3.0}$ or lower.

\noindent \textbf{Comparisons with ACADO:} We compared our optimizer with ACADO that uses multiple-shooting along with SQP for solving optimal control problems. We did the comparison in trajectory optimization setting. That is, the optimization iteration is continued till a feasible and a low cost solution is obtained. Alternately, ACADO is also popularly used in so called Real-Time iteration based Model Predictive Control (MPC) setting, where only one optimization iteration is performed \cite{multiple_shoot_fwv_1}. The FWV moves with the sub-optimal solution with the belief that subsequent re-planning will improve the solution. Our chosen trajectory optimization setup allows us to benchmark the full capability of our optimizer. Moreover, often the MPC itself needs to be warm-started with close to optimal solution in the first iteration and for which we need an fast trajectory optimizer.

\begin{figure*}
     \centering
         \includegraphics[scale=0.38]{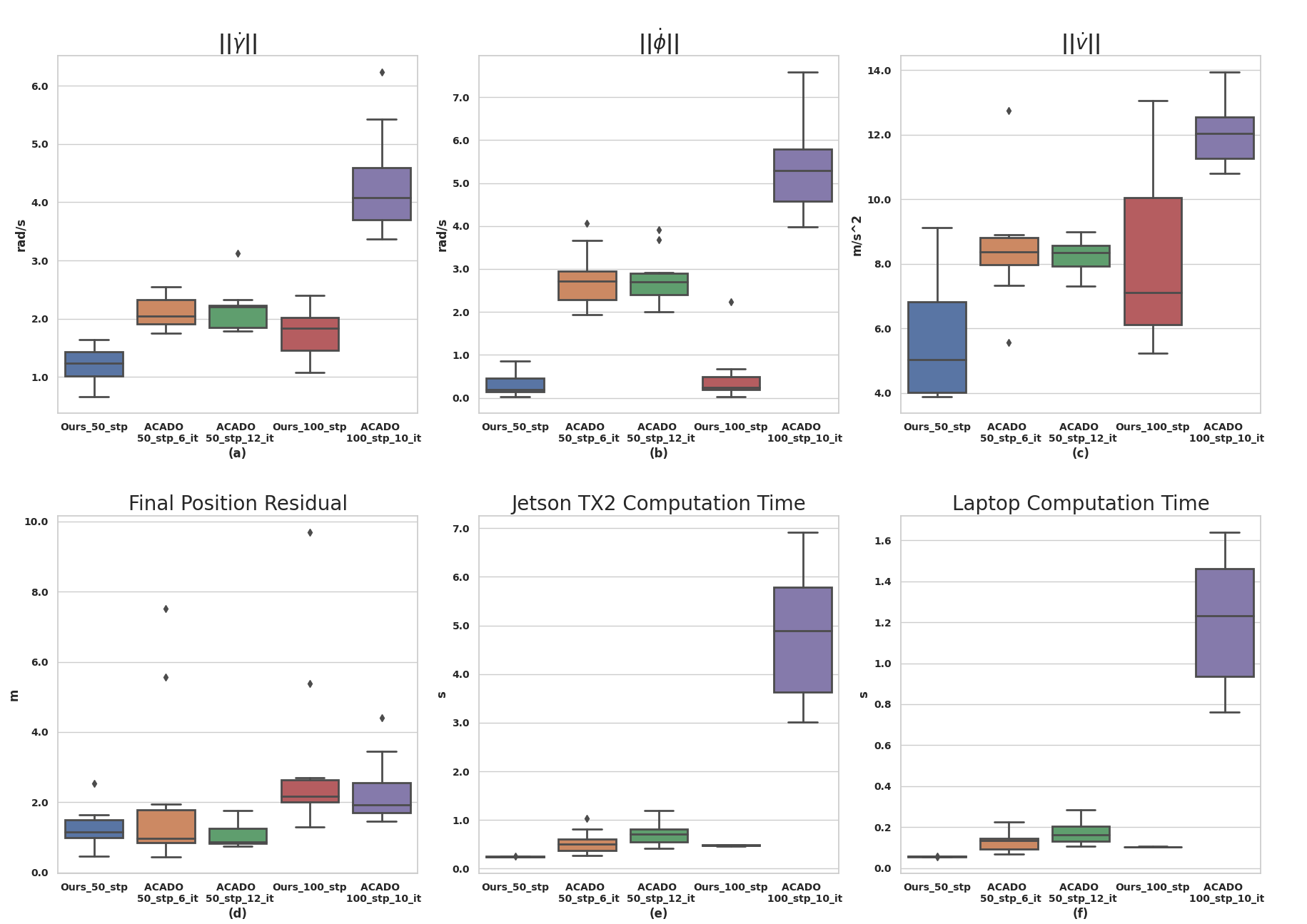}
          \caption{Convergence Validation: Fig. (a),(b), (c) shows that our optimizer leads to trajectories with smaller control input norms as compared to ACADO.  Fig. (d) quantifies the statistics of how far our optimizer and ACADO computed trajectory is from the goal position. Fig.(e),(f) shows that our optimizer is faster than ACADO on both laptops and Jetson TX2. }
        \label{bench_acado}
        \vspace{-0.6cm}
\end{figure*}

ACADO is provided with the problem description in the continuous form (\ref{cost_standard})-(\ref{coll_nonhol_standar}). We made our best efforts to tune ACADO's performance. This involved tuning the trade-off between control and the goal reaching costs and providing the travel time heuristic that our optimizer uses to ACADO at no extra computation overhead. For faster computation, we used ACADO in so-called code-generation setup that generates custom C codes for a given problem. However, this setup also requires us to manually try different iteration limits in order to ensure we reach the  constraint tolerances comparable to our optimizer. The following metrics were used for quantitative comparisons. 

\begin{itemize}
    \item Norm of the control inputs: We stacked $\dot{\gamma}(t), \dot{\phi}(t), \dot{v}(t)$ at different time instants and computed the resulting norm. Our optimizer does not compute these terms explicitly but instead obtain them from finite difference of $\gamma(t), v(t), \psi(t)$.
    
    \item Final position residual: difference between desired goal point and actual end point of the trajectory.
    
    \item Computation time: We evaluated computation times on both Jetson TX2 and a regular Laptop. 
\end{itemize}

\noindent {$\Vert \dot{\boldsymbol{\gamma}} \Vert $ \textbf{comparison, Fig.\ref{bench_acado} (a)}:} For the benchmarks with 50 planning steps, the median values obtained by our optimizer is $1.232$ $rad/s$ while that obtained with ACADO with 6 iteration limit is $2.04$ $rad/s$. Thus, our optimizer achieves a median reduction of $41\%$ over ACADO. The worst and best (upper and lower whiskers) case value obtained by our optimizer across all configurations is $1.642$ $rad/s$ and $0.66$ $rad/s$ respectively. The same values for ACADO stands at  $2.55$ $rad/s$ and $1.75$ $rad/s$ respectively. Thus, our worst case performance is around $4\%$ better than the ACADO's best performance if the iteration limit is kept at 6. The worst case performance becomes slightly better for ACADO if we increase the iteration limit to 12. However, this comes at the expense of a higher median value of $2.2$ $rad/s$. For ACADO with 12 iteration limit, we also encounter a single outlier at $3.2$ $rad/s$. 

For benchmarks with 100 planning steps, our optimizer achieves even greater improvement over ACADO. The median value of $\Vert \dot{\boldsymbol{\gamma}} \Vert $ resulting from our optimizer is $1.82$ $rad/s$, while that achieved with ACADO is $2.2$ times higher at $4.08$ $rad/s$. In the worst case, our optimizer ensures a reduction of $30\%$ over values achieved with ACADO.

\noindent {$\Vert \dot{\boldsymbol{\phi}} \Vert  $ \textbf{comparison, Fig.\ref{bench_acado} (b)}:} The trends here are even more skewed in the favor of our optimizer. For 50 step benchmarks, our optimizer achieves median values of $0.195$ $rad/s$. In contrast ACADO with 6 iteration limit, is more than $13$ times higher at $2.72$ $rad/s$. Moreover, if we compare the maximum value resulting from our optimizer ($0.86$ $rad/s$) to the minimum value achieved with ACADO ($1.93$ $rad/s$), we obtain a worst-case improvement of $100\%$.   
For this metric, we did not observe any statistical gain by increasing the ACADO iteration limit to $12$. 

On moving to benchmarks with $100$ steps, the $\Vert \dot{\boldsymbol{\phi}} \Vert  $ achieved by our optimizer did not change much. However, values achieved by ACADO increased substantially. The median and worst difference between the two approaches were $20$ and $4$ times respectively.

\noindent {$\Vert \dot{\textbf{v}} \Vert  $ \textbf{comparison, Fig.\ref{bench_acado} (c)}:} Our optimizer leads to trajectories that require lower forward acceleration inputs as compared to those computed by ACADO. For benchmarks with 50 planning steps, our optimizer achieves median values of $5.02$ $m/s^2$. In comparison, ACADO with 6 iteration limit leads to a median value of $8.36$ $m/s^2$. We do not observe any improvement by increasing the iteration limit to 12. However, interestingly, our optimizer shows a large variance in the forward acceleration values. Thus, the worst case achieved by our optimizer ($9.13$ $m/s^2$) is marginally higher than that ensured by ACADO ($8.9$ $m/s^2$). On the other end of the spectrum, the best case values achieved by our optimizer is 1.8 times lower than ACADO's. In 100 planning step benchmarks, forward acceleration median values of our optimizer are 1.7 times lower than ACADO's.

\noindent {\textbf{Final position residual comparison, Fig.\ref{bench_acado} (d)}:} We observed very interesting trends for this metric. ACADO's consistently achieves marginally lower median residual values than our optimizer. For benchmarks with 50 planning steps, our optimizer's median residual is $1.14$ $m$ while that of ACADO's is $0.979$ $m$. ACADO's median residual decreases to $0.86$ $m$ if we increase the iteration limit to 12. In 100 planning step benchmarks, the median values observed for our optimizer and ACADO are $2.16$ $m$ and $1.97$ $m$ respectively. For this metric, we also observed, large variance in the data for both the approaches, making worst-case prediction difficult. 

\noindent {\textbf{Computation Time comparison, Fig.\ref{bench_acado} (e), (f)}:} These figures represent the most important result of the paper. The median computation time of our optimizer on Jetson TX2 for benchmarks with 50 planning steps is $0.24s$. This is $2.24/3$ times lower than ACADO's computation time with 6/12 iteration limit. Additionally, our optimizer's worst-case computation time is two times lower than the best possible time that ACADO takes for 12 iterations. The difference in computation time becomes even more stark when we consider benchmarks with 100 planning steps. Our optimizer's median computation time in this setting is $0.47s$, almost one order of magnitude lower than ACADO's $4.85s$. Even at the worst-case our optimizer is 6 times faster than ACADO. The trends (median/worst-case) in computation time remains same on a i7-8750, 32 GB RAM Laptop with computation times reducing for both our optimizer and ACADO. Notably, our optimizer's median run times reduce to  just $0.05s$.

Our optimizer's computational efficiency stems from how we handle collision-avoidance and integrate them with kinematic constraints within an alternating minimization framework. Our preliminary evaluation showed that incorporating just our collision avoidance model in ACADO solver led to inferior performance. The reason being, on the surface, our collision avoidance constraints are more complex. Only when we group the optimization variables in specific blocks and adopt the AM technique, we get the convex unconstrained QP structure in each minimization step. In contrast, the SQP approach based on Taylor Series expansion that ACADO uses cannot identify or leverage the niche computational structures. 

\vspace{-0.3cm}

\section{Conclusions and Future Work}
We presented an embedded hardware appropriate 3D trajectory optimizer for FWVs. We showed for the first time how this problem has niche computational structures that we can leverage for faster and reliable computations. We substantially outperformed the state-of-the-art implementation of SQP in terms of control costs and computation time. Our optimizer scales linearly with the number of planning steps. Thus, we can reduce the computation time from $0.25 s$ (observed for 50 steps) to real-time levels on Jetson TX2 by restricting the planning horizon to 10.

One drawback of our optimizer is that it has sub-linear convergence, which is typical for any AM/BI/ADMM based approach \cite{boyd_admm}. Intuitively, this means that moderate constraint residuals ($10^{-3.0}$) can be achieved quickly, but substantially more iterations are needed to reach residuals of the order of ($10^{-6.0}$). However, in practice, the mentioned moderate residuals prove sufficient.

In future, we will be extending our optimizer to multiple FWVs. We also aim to explore if multi-convex structures also exists in the dynamics of FWVs. We also aim to develop a receding horizon variant of our optimizer and perform extensive hardware experiments to further analyze the capabilities of our optimizer.

\vspace{-0.1cm}

\bibliographystyle{IEEEtran}  
\bibliography{iros_2021_fwv}

\end{document}